\pdfoutput=1

\documentclass[11pt]{article}

\usepackage{naacl2021}
\usepackage{placeins}
\usepackage{times}
\usepackage{latexsym}
\usepackage{multirow}
\usepackage{subcaption}
\usepackage{graphicx, caption}
\usepackage[T1]{fontenc}
\usepackage[utf8]{inputenc}
\usepackage{microtype}
\usepackage{amsmath}
\usepackage{amssymb}
\usepackage{url}
\usepackage[export]{adjustbox}

\title{Multi-Modal Image Captioning for the Visually Impaired}

\author{Hiba Ahsan\thanks{\hspace{2mm}Equal contribution}\qquad Nikita Bhalla\footnotemark[1] \qquad
\qquad Daivat Bhatt\footnotemark[1] \qquad Kaivankumar Shah\footnotemark[1] \\\texttt{\{hahsan,nbhalla,dbhatt,kaivankumars\}@umass.edu}\\University of Massachusetts Amherst}

\begin{document}
\maketitle
\begin{abstract}

One of the ways blind people understand their surroundings is by clicking images and relying on descriptions generated by image captioning systems. Current work on captioning images for the visually impaired do not use the textual data present in the image when generating captions. This problem is critical as many visual scenes contain text. Moreover, up to 21\% of the questions asked by blind people about the images they click pertain to the text present in them \cite{10.1145/1866029.1866080}. In this work, we propose altering AoANet, a state-of-the-art image captioning model, to leverage the text detected in the image as an input feature. In addition, we use a pointer-generator mechanism to copy the detected text to the caption when tokens need to be reproduced accurately. Our model outperforms AoANet on the benchmark dataset VizWiz, giving a 35\% and 16.2\% performance improvement on CIDEr and SPICE scores, respectively.
\end{abstract}. 

\section{Introduction}
\begin{figure}[htp]
\centering
\begin{subfigure}{.22\textwidth}
  \centering
  \includegraphics[width=\linewidth]{figs/VizWiz_val_00003999.jpg}
  \caption{\textbf{Model}: a bottle of water is on top of a table \\ \textbf{Ground Truth:} a clear plastic bottle of Springbourne brand spring water}
  \label{fig:Intro1}
\end{subfigure}%
\hspace{0.5cm}
\begin{subfigure}{.22\textwidth}
  \centering
  \includegraphics[width=\linewidth]{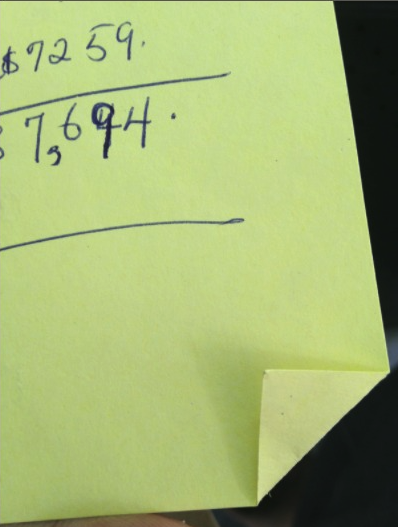}
  \caption{\textbf{Model}: A piece of paper with text on it \newline \textbf{Ground Truth}: In a yellow paper written as 7259 and totally as 7694}
  \label{fig:Intro2}
\end{subfigure}
\label{fig:test}
\end{figure}

Image Captioning as a service has helped people with visual impairments to learn about images they take and to make sense of images they encounter in digital environments. Applications such as \cite{TTS} allow the visually impaired to take photos of their surroundings and upload them to get descriptions of the photos. Such applications leverage a human-in-the-loop approach to generate descriptions. In order to bypass the dependency on a human, there is a need to automate the image captioning process. Unfortunately, the current state-of-the-art (SOTA) image captioning models are built using large, publicly available, crowd-sourced datasets which have been collected and created in a contrived setting. Thus, these models perform poorly on images clicked by blind people largely because the images clicked by blind people differ dramatically from the images present in the datasets. To encourage solving this problem, \citet{gurari2020captioning} released the VizWiz dataset, a dataset comprising of images taken by the blind. Current work on captioning images for the blind do not use the text detected in the image when generating captions (Figures \ref{fig:Intro1} and \ref{fig:Intro2} show two images from the VizWiz dataset that contain text). The problem is critical as many visual scenes contain text and up to 21\% of the questions asked by blind people about the images clicked by them pertain to the text present in them. This makes it more important to improvise the models to focus on objects as well as the text in the images. 

With the availability of large labelled corpora, image captioning and reading scene text (OCR) have seen a steady increase in performance. 
However, traditional image captioning models focus only on the visual objects when generating captions and fail to recognize and reason about the text in the scene. This calls for incorporating OCR tokens into the caption generation process. The task is challenging since unlike conventional vocabulary tokens which depend on the text before them and therefore can be inferred, OCR tokens often cannot be predicted from the context and therefore represent independent entities. Predicting a token from vocabulary and selecting an OCR token from the scene are two rather different tasks which have to be seamlessly combined to tackle this task. 

In this work, we build a model to caption images for the blind by leveraging the text detected in the images in addition to visual features. We alter AoANet, a SOTA image captioning model to consume embeddings of tokens detected in the image using Optical Character Recognition (OCR). In many cases, OCR tokens such as entity names or dates need to be reproduced exactly as they are in the caption. To aid this copying process, we employ a pointer-generator mechanism. Our contributions are 1) We build an image captioning model for the blind that specifically leverages text detected in the image. 2) We use a pointer-generator mechanism when generating captions to copy the detected text when needed. 

\section{Related Work}
Automated image captioning has seen a significant amount of recent work. The task is typically handled using an encoder-decoder framework; image-related features are fed to the encoder and the decoder generates the caption \cite{aneja2018convolutional, yao2018exploring, cornia2018paying}. Language modeling based approaches have also been explored for image captioning \cite{kiros2014unifying, devlin2015language}. Apart from the architecture, image captioning approaches are also diverse in terms of the features used. Visual-based image captioning models exploit features generated from images. Multi-modal image captioning approaches exploit other modes of features in addition to image-based features such as candidate captions and text detected in images \cite{wang2020multimodal, hu2020iterative}.

The task we address deals with captioning images specifically for the blind. This is different from traditional image captioning due to the authenticity of the dataset compared to popular, synthetic ones such as MS-COCO \cite{DBLP:journals/corr/ChenFLVGDZ15} and Flickr30k \cite{plummer2015flickr30k} . The task is relatively less explored. Previous works have solved the problem using human-in-the-loop approaches \cite{Aira, BeSpec, TTS} as well as automated ones \cite{MS, FB}. A particular challenge in this area has been the lack of an authentic dataset of photos taken by the blind. To address the issue, \citet{gurari2020captioning} created VizWiz-Captions, a dataset that consists of descriptions of images taken by people who are blind. In addition, they analyzed how the SOTA image captioning algorithms performed on this dataset. Concurrent to our work, \citet{dognin2020image} created a multi-modal transformer that consumes ResNext based visual features, object detection-based textual features and OCR-based textual features. Our work differs from this approach in the following ways: we use AoANet as our captioning model and do not account for rotation invariance during OCR detection. We use BERT to generate embeddings of the OCR tokens instead of fastText. Since we use bottom-up image feature vectors extracted using a pre-trained Faster-RCNN, we do not use object detection-based textual features. Similarly, since the Faster-RCNN is initialized with ResNet-101 pre-trained for classification, we do not explicitly use classification-based features such as those generated by ResNext.

We explored copy mechanism in our work to aid copying over OCR tokens from the image to the caption. Copy mechanism has been typically employed in textual sequence-to-sequence learning for tasks such as summarization \cite{see2017get, gu2016incorporating}. It has also been used in image captioning to aid learning novel objects \cite{yao2017incorporating, li2019pointing}. Also, \citet{Sidorov2020TextCapsAD} introduced an M4C model that recognizes text, relates it to its visual context, and decides what part of the text to copy or paraphrase, requiring spatial, semantic, and visual reasoning between multiple text tokens and visual entities such as objects.
\section{Dataset} 
The Vizwiz Captions dataset \cite{gurari2020captioning} consists of over $39,000$ images originating from people who are blind that are each paired with five captions. The dataset consists of $23,431$ training images, $7,750$ validation images and $8,000$ test images. The average length of a caption in the train set and the validation set was 11. We refer readers to the \href{https://ivc.ischool.utexas.edu/VizWiz_visualization/view_dataset.php}{VizWiz Dataset Browser} \cite{vizwiz_browser} as well as the original paper by \citet{gurari2020captioning} for more details about the dataset.  

\section{Approach}
\label{approach}
We employ AoANet as our baseline model. AoANet extends the conventional attention mechanism to account for the relevance of the attention results with respect to the query. 
An attention module $f_{att}(Q,K,V)$ operates on queries $Q$, keys $K$ and values $V$. It measures the similarities between $Q$ and $K$ and using the similarity scores to compute a weighted average over $V$.
\begin{align}
a_{i,j} = f_{sim}(q_i,k_j)&,  \alpha=\frac{e^{a_{i,j}}}{\sum_je^{a_{i,j}}}\\
\hat{v}_i &= \sum_j \alpha_{i,j}v_{i,j}\\
f_{sim}(q_i,k_j)&=\text{softmax}(\frac{q_ik_j^T}{\sqrt[]{D}})v_i
\end{align}
where $q_i\in Q$ is the $i^{th}$ query, $k_j \in K$ and $v_j \in V$  are the $j^{th}$  key/value pair, $f_{sim}$ is the similarity function, $D$ is the dimension of $q_i$ and $\hat{v}_i$ is the attended vector for query $q_i$.

The AoANet model introduces a module AoA which measures the relevance between the attention result and the query. The AoA module generates an "information vector", $i$, and an "attention gate", $g$, both of which are obtained via separate linear transformations, conditioned on the attention result and the query:
\begin{align}
    i &= W^i_{q}q + W^i_{v}\hat{v} + b^i\\
    g &= \sigma(W^g_{q}q + W^g_{v}\hat{v} + b^g)
\end{align}
where $W^i_q, W^i_{v}, b^i, W^g_{q}, W^g_{v}, b^g$ are parameters. AoA module then adds another attention by applying the attention gate to the information vector to obtain the attended information $\hat{i}$.
\begin{align}
    \hat{i}=g\odot i
\end{align}
The AoA module can thus be formulated as:
\begin{multline}
    AoA(f_{att},Q,K,V) = \sigma(W^g_{q}Q + W^g_{v}f_{att}(Q,\\K,V)+ b^g) \odot (W^i_{q}Q + W^i_{v}f_{att}(Q,K,V)+ b^i)
\end{multline}
The AoA module is applied to both the encoder and decoder. The model is trained by minimizing the cross-entropy loss:
 \begin{align}
     L(\theta) = -\sum^T_{t=1}\log(p_\theta(y^{*}_t|y^{*}_{1:t-1}))
 \end{align}
where $y^{*}_{1:T}$ is the ground truth sequence. We refer readers to the original work \cite{huang2019attention} for more details. We altered AoANet using two approaches described next.

\subsection{Extending Feature Set with OCR Token Embeddings}
\label{extended}
Our first extension to the model was to increase the vocabulary by incorporating OCR tokens. 
We use an off-the-shelf text detector available - Google Cloud Platform's vision API \cite{Google}. After extracting OCR tokens for each image using the API, we use a standard stopwords list\footnote{\href{Stopwords: }{https://gist.github.com/sebleier/554280}} as part of necessary pre-processing. 
We use this API to detect text in an image and then generate an embedding for each OCR token that we detect using a pre-trained base, uncased BERT \cite{devlin-etal-2019-bert} model. 
The image and text features are fed together into the AoANet model. We expect the BERT embeddings to help the model direct its attention towards the textual component of the image. Although we also experiment with a pointer-generator mechanism explained in Section \ref{copying}, we wanted to leverage the model's inbuilt attention mechanism that currently performs as a state of the art model and guide it towards using these OCR tokens.

\begin{figure}[h]
    \centering
    \includegraphics[width=\columnwidth]{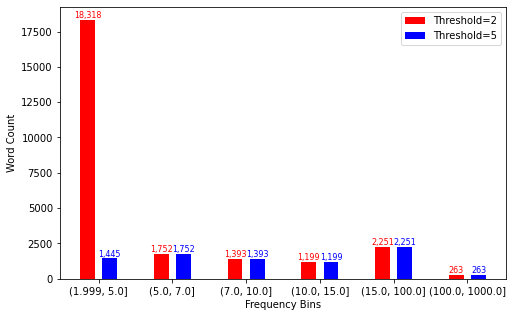}
    \caption{Word counts and frequency bins for threshold = 2 and threshold = 5}
    \label{fig:ocr_stats_combined}
\end{figure}

Once the OCR tokens were detected, we conducted two different experiments with varying sizes of thresholds. We first put a count threshold of 5 i.e. we only add words to the vocabulary which occur 5 or more times. 
With this threshold, the total words added were $4,555$. We then put a count threshold of 2. With such a low threshold, we expect a lot of noise to be present in the OCR tokens vocabulary - half-detected text, words in a different language, or words that do not make sense. With this threshold, the total words added were $19,781$. A quantitative analysis of the OCR tokens detected and their frequency is shown in Figure \ref{fig:ocr_stats_combined}.




\subsection{Copying OCR Tokens via Pointing}
\label{copying}
In sequence-to-sequence learning, there is often a need to copy certain segments from the input sequence to the output sequence as they are. This can be useful when sub-sequences such as entity names or dates are involved. Instead of heavily relying on meaning, creating an explicit channel to aid copying of such sub-sequences has been shown to be effective \cite{gu2016incorporating}. 

In this approach, in addition to augmenting the input feature set with OCR token embeddings, we employ the pointer-generator mechanism \cite{see2017get} to copy OCR tokens to the caption when needed. The decoder then becomes a hybrid that is able to copy OCR tokens via pointing as well as generate words from the fixed vocabulary. A soft-switch is used to choose between the two modes. The switching is dictated by \textit{generation probability}, $p_{gen}$, calculated at each time-step, $t$, as follows:
\begin{align}
    p_{gen}=\sigma(w_h^Tc_t + w_s^Th_t + w_x^Tx_t + b_{ptr})
\end{align}
where $\sigma$ is the sigmoid function and $w_h, w_s, w_x \text{ and } b_{ptr}$ are learnable parameters. $c_t$ is the context vector, $h_t$ is the decoder hidden state and $x_t$ is the input embedding at time $t$ in the decoder.
At each step, $p_{gen}$ determines whether a word has to be generated using the fixed vocabulary or to copy an OCR token using the attention distribution at time $t$. Let \textit{extended vocabulary} denote a union of the fixed vocabulary and the OCR words. The probability distribution over the \textit{extended vocabulary} is given as:
\begin{align}
    P(w)=p_{gen}P_{vocab}(w)+(1-p_{gen})\sum_{i:w_i=w}a_i^t
\end{align}
$P_{vocab}$ is the probability of $w$ using the fixed vocabulary and $a$ is the attention distribution. If $w$ does not appear in the fixed vocabulary, then $P_{vocab}$ is zero. If $w$ is not an OCR word, then $\sum_{i:w_i=w}a_i^t$ is zero. \\

\section{Experiments}
In our experiments, we alter AoANet as per the approaches described in Section \ref{approach} and compare these with the baseline model. AoANet-E refers to AoANet altered as per the approach described in Section \ref{extended}. To observe the impact of the number of OCR words added to the extended vocabulary, we train two Extended variants: (1) E5: Only OCR words with frequency greater than or equal to 5. (2) E2: Only OCR words that occur with frequency greater than or equal to 2. AoANet-P refers to AoANet altered as per the approach described in Section \ref{copying}. The extended vocabulary consists of OCR words that occur with frequency greater than or equal to 2.

We use the code\footnote{\href{AoANet github repo: }{https://github.com/husthuaan/AoANet}} released by the authors of AoANet to train the model. We cloned the repository and made changes to extend the feature set and the vocabulary using OCR tokens as well as to incorporate the copy mechanism during decoding \footnote{\href{Altered AoANet github repo: }{https://github.com/hiba008/AlteredAoA}}. We train our models on a Google Cloud VM instance with 1 Tesla K80 GPU. 
Like the original work, we use a Faster-RCNN \cite{ren2015faster} model pre-trained on ImageNet \cite{deng2009imagenet} and Visual Genome \cite{krishna2017visual} to extract bottom-up feature vectors of images. The OCR token embeddings are extracted using a pre-trained base, uncased BERT model. The AoANet models are trained using the Adam optimizer and a learning rate of $2e-5$ annealed by 0.8 every 3 epochs as recommended in \citet{huang2019attention}. The baseline AoANet is trained for 10 epochs while AoANet-E and AoANet-P are trained for 15 epochs.

\begin{table*}[]
\centering
\begin{tabular}{|c|c|c|c|c|c|c|c|c|}
\hline
\multirow{2}{*}{\textbf{Model}} & \multicolumn{4}{c|}{\textbf{Validation Scores}} & \multicolumn{4}{c|}{\textbf{Test Scores}}  \\ \cline{2-9} 
& BLEU-4   & ROUGE\_L  & SPICE  & CIDEr  & BLEU-4 & ROUGE\_L & SPICE & CIDEr \\ \hline\hline
AoANet& 21.4&43.8&11.1&40.0&19.5&43.1&12.2&40.8\\ \hline
AoANet-E5              &  21.4        & 43.6          &  10.8      & 41.4        &  19.8       & 42.9          & 11.9       & 40.2       \\ \hline
AoANet-E2              & \textbf{24.3}          &  \textbf{46.1}         & \textbf{12.9}      & \textbf{54.1}        &  \textbf{22.3}      & \textbf{45.0}         &  \textbf{14.1}     &  \textbf{53.8}    \\ \hline
AoANet-P&21.6&43.6&11.5&46.1&19.9&42.7&12.8&45.4      \\ \hline
\end{tabular}%
\caption{Validation and Test scores for AoANet, AoANet-E5 (extended vocabulary variant with OCR frequency threshold as 5), AoANet-E2 (extended vocabulary variant with OCR frequency threshold as 2) and AoANet-P (pointer-generator variant).} 
\label{metric_scores}
\end{table*}



\section{Results}
We show quantitative metrics for each of the models that we experimented with in Table \ref{metric_scores}.
We show qualitative results where we compare captions generated by different models in Table \ref{aoa-dataset-examples}. 
Note that none of the models were pre-trained on the MS-COCO dataset as \citet{gurari2020captioning} have done as part of their experimenting process.


We compare different models and find that merely extending the vocabulary helps to improve model performance on the dataset. We see that the AoAnet-E5 matches the validation scores for AoANet but we see an improvement in the CIDEr score. Moreover, we see a massive improvement in validation and test CIDEr scores for AoANet-E2. Similarly, we see a gain in the other metrics too. This goes to show that the BERT embeddings generated for each OCR token for the images do provide an important context to the task of generating captions. Moreover, we see the AoANet-P scores, where we use pointer-generator to copy OCR tokens after extending the vocabulary also perform better than our baseline AoANet model. This goes to show that an OCR copy mechanism is an essential task in generating image captions. Intuitively, it makes sense because we would expect to humans to use these words while generating lengthy captions ourselves.

We feel that \textit{top-k} sampling is a worthwhile direction of thought especially when we would like some variety in the captions. Beam-search is prone to preferring shorter captions, as the probability values for longer captions accumulates smaller values as discussed by \citet{topkpaper}.

\section{Error Analysis}
\label{error}
Although there have been concerns about the robustness of the GCP API towards noise \cite{robustness-vision}, we focused our attention on the model's captioning performance and on the pointer-generator mechanism. We agree that the API's performance might hinder the quality of the captions generated but we expected it to not have a large enough impact. 

We first look at how the Extended variants compare with the baseline. We observe that adding text-based features to the feature set imparts useful information to the model. In \ref{fig:23693}, AoANet perceives the card as a box of food. Addition of text features enables AoANet-E5 to perceive it as a box with black text. While not entirely correct, it is an improvement over the baseline. The alteration also encourages it to be more specific. When the model is unable to find the token that entails specificity, it resorts to producing UNK. Extending the vocabulary to accommodate more OCR words helps address this problem. In image \ref{fig:25800}, baseline AoANet is unable to recognize that the bottle is a supplements bottle. AoANet-E5 attempts to be specific but since 'dietary' and 'supplement' are not present in the extended vocabulary, it outputs UNK. AoANet-E2 outputs a much better caption. We see a similar pattern in \ref{fig:25901}.

We now look at how the Pointer variant performs compared to the baseline and the Extended variant. Incorporating copy mechanism helps the Pointer variant in copying over OCR tokens to the caption. AoANet-P is able to copy over `oats' and `almonds' in \ref{fig:24776} and the token `rewards' in \ref{fig:23824}. But the model is prone to copying tokens multiple times as seen in images \ref{fig:25800}  and \ref{fig:25774}. This is referred to as repetition which is a common problem in sequence-to-sequence models \cite{tu2016modeling} as well as in pointer generator networks. Coverage mechanism \cite{tu2016modeling, see2017get} is used to handle this and we wish to explore this in the future.

\section{Conclusion}
In this work, we propose a pointer-generator based image captioning model that deals specifically with images taken by people with visual disabilities. Our alteration of AoANet shows significant improvement on the VizWiz dataset compared to the baseline. As stated in Section \ref{error}, we would like to explore coverage mechanism in the future. 
\citet{dognin2020image} recently discussed their winning entry to the VizWiz Grand Challenge. In addition, \citet{Sidorov2020TextCapsAD} introduced a model  that has shown  to  gain  significant  performance  improvement by using OCR tokens. We intend to compare our model with these and  improve our work based on the observations made. 

\section{Acknowledgements}
The authors would like to thank Mohit Iyyer and Kalpesh Krishna at the University of Massachusetts, Amherst for their invaluable guidance and suggestions. The authors would also like to thank the University of Massachusetts, Amherst for providing the necessary resources throughout the course of this project.
\begin{table*}[ht]
\small
\begin{tabular}{|p{0.2\linewidth}|p{0.8\linewidth}|}
\hline
Image&Captions\\
\hline
\multirow{10}{\linewidth}
{\begin{subfigure}{0.2\textwidth}
\centering
\includegraphics[width=80pt, height=70pt]{figs/VizWiz_val_00000262.jpg}
\caption{}
\label{fig:23693}
\end{subfigure}}&\textbf{AoANet}: the back of a box of food that is yellow \\
&\textbf{AoANet-E5}: the back of a yellow box with black text\\
&\textbf{AoANet-E2}: the back of a card with a barcode on it\\
&\textbf{AoANet-P}: the back of a UNK UNK card\\
&\textbf{GT1}: The back of an EBT card that is placed on a black surface.\\
&\textbf{GT2}: The back of a California EBT debit card.\\
&\textbf{GT3}: A yellow EBT card on a dark fabric surface.\\
&\textbf{GT4}: The backside of a beige EBT card with a magnetic strip.\\
&\textbf{GT5}: back of yellow Quest card with black text on it and a white empty signature box\\
&\\
\hline

\multirow{10}{\linewidth}
{\begin{subfigure}{0.2\textwidth}
\centering
\includegraphics[width=80pt, height=70pt]{figs/VizWiz_val_00002369.jpg}
\caption{}
\label{fig:25800}
\end{subfigure}} &\textbf{AoANet}:a person is holding a bottle of seasoning \\
&\textbf{AoANet-E5}: a person is holding a bottle of UNK\\
&\textbf{AoANet-E2}: a person is holding a bottle of dietary supplement\\
&\textbf{AoANet-P}: a person is holding a bottle of super tablets tablets tablets tablets tablets tablets\\
&\textbf{GT1}: A bottle of Nature's Blend Vitamin D3 2000 IU with 100 tablets.\\
&\textbf{GT2}: bottle of Nature's Blend brand vitamin D3 tablets, 100 count, 2000 IU per tab\\
&\textbf{GT3}: A hand is holding a container of vitamin D.\\
&\textbf{GT4}: Someone is holding a black bottle with a yellow lid.\\
&\textbf{GT5}: A person's hand holds a bottle of Vitamin D3 tablets.\\
&\\
\hline
\multirow{10}{\linewidth}
{\begin{subfigure}{0.2\textwidth}
\centering
\includegraphics[width=80pt, height=70pt]{figs/VizWiz_val_00002470.jpg}
\caption{}
\label{fig:25901}
\end{subfigure}}&\textbf{AoANet}: a a green bottle with a green and white label \\
&\textbf{AoANet-E5}: a green bottle of UNK UNK UNK UNK\\
&\textbf{AoANet-E2}: a bottle of body lotion is on a table\\
&\textbf{AoANet-P}: a bottle of vanilla lotion is sitting on a table\\
&\textbf{GT1}: A container of vanilla bean body lotion is on a white table.\\
&\textbf{GT2}: A bottle of body lotion sits on top of a white table\\
&\textbf{GT3}: a plastic bottle of vanilla bean body lotion from bath and body works\\
&\textbf{GT4}: A bottle of body lotion that says Noel on it sitting on a table with a phone behind it and other items around it.\\
&\textbf{GT5}: A body lotion bottle is on top of table with several papers behind it and a set of keys in the background.\\
&\\
\hline
\multirow{10}{\linewidth}
{\begin{subfigure}{0.2\textwidth}
\centering
\includegraphics[width=80pt, height=70pt]{figs/VizWiz_val_00001345.jpg}
\caption{}
\label{fig:24776}
\end{subfigure}}&\textbf{AoANet}: a box of frozen dinner is on top of a table \\
&\textbf{AoANet-E5}: a box of UNK 's UNK brand UNK UNK\\
&\textbf{AoANet-E2}: a box of granola granola granola granola bars\\
&\textbf{AoANet-P}: a box of oats 's almond almond bars\\
&\textbf{GT1}: A box of nature valley roasted almond crunchy bars is on a table.\\
&\textbf{GT2}: A box of granola bars sitting on a floral cloth near a wooden object.\\
&\textbf{GT3}: A granola bar box sits on a table cloth with other items.\\
&\textbf{GT4}: Green box with roasted almond granola bar  place tablecloth with flower prints.\\
&\textbf{GT5}: A package of granola bars is lying on top of a table.\\
&\\

\hline
\multirow{10}{\linewidth}
{\begin{subfigure}{0.2\textwidth}
\centering
\includegraphics[width=80pt, height=70pt]{figs/VizWiz_val_00000393.jpg}
\caption{}
\label{fig:23824}
\end{subfigure}}&\textbf{AoANet}: a hand holding a box of chocolate 's brand \\
&\textbf{AoANet-E5}: a person is holding a package of food\\
&\textbf{AoANet-E2}: a hand holding a card with a number on it\\
&\textbf{AoANet-P}: a person is holding a box of rewards card\\
&\textbf{GT1}: Appears to  be a picture of a reward card\\
&\textbf{GT2}: A plastic card that says speedy rewards membership card.\\
&\textbf{GT3}: A Speedy Rewards membership card with a large gold star displayed on it.\\
&\textbf{GT4}: a human hold some cards like credit cards and reward cards\\
&\textbf{GT5}: Rewards membership card from the Speedway chain of stores.\\
&\\

\hline
\multirow{10}{\linewidth}
{\begin{subfigure}{0.2\textwidth}
\centering
\includegraphics[width=80pt, height=70pt]{figs/VizWiz_val_00002343.jpg}
\caption{}
\label{fig:25774}
\end{subfigure}}&\textbf{AoANet}: a bottle of water is on top of a table \\
&\textbf{AoANet-E5}:  a bottle of water is on top of a table\\
&\textbf{AoANet-E2}: a bottle of vanilla vanilla coffee mate creamer\\
&\textbf{AoANet-P}: a bottle of vanilla vanilla vanilla vanilla vanilla\\
&\textbf{GT1}: A bottle of coffee creamer has a plastic flip top cap that can also be twisted off.\\
&\textbf{GT2}: A blue bottle of coffee creamer is sitting on a counter top next to a  black cup.\\
&\textbf{GT3}: A container of Coffee Mate French Vanilla showing part of the front and part of the back.\\
&\textbf{GT4}: A bottle of French vanilla coffee creamer sits in front of a mug on the table.\\
&\textbf{GT5}: A bottle of creamer is on top of a table.\\
&\\
\hline
\end{tabular}
\caption{\label{aoa-dataset-examples} Examples of captions generated by AoANet, AoANet-E5 (extended vocabulary variant with OCR frequency threshold as 5), AoANet-E2 (extended vocabulary variant with OCR frequency threshold as 2) and AoANet-P (pointer-generator variant) for validation set images along with their respective ground truth captions.
}
\end{table*}
\FloatBarrier

\bibliographystyle{acl_natbib}
\bibliography{custom}




\end{document}